\documentclass{article}

\usepackage{arxiv}
\usepackage[utf8]{inputenc} % allow utf-8 input
\usepackage[T1]{fontenc}    % use 8-bit T1 fonts
\usepackage{hyperref}       % hyperlinks
\usepackage{url}            % simple URL typesetting
\usepackage{booktabs}       % professional-quality tables
\usepackage{amsfonts}       % blackboard math symbols
\usepackage{nicefrac}       % compact symbols for 1/2, etc.
\usepackage{microtype}      % microtypography
\usepackage{lipsum}
\usepackage{graphicx}
\usepackage{multicol}
\graphicspath{ {./img/} }

\title{Automatic Discourse Segmentation:\\an evaluation in French}

\author{
  Rémy Saksik\\
  Research engineer\\
  Thales DIS (Digital Identity and Security)\\
  Singapore\\
  \texttt{saksikremy@gmail.com} \\
   \And
  Alejandro Molina-Villegas\\
  CONACyT – CentroGeo\\
  Yucatán, Mexico\\
  \texttt{amolina@centrogeo.edu.mx} \\
  \And
  Andréa Carneiro Linhares\\
   Universidade Fédérale do Cear\'a\\
  Brazil\\
  \texttt{andreaclinhares@gmail.com} \\
   \AND
   Juan-Manuel Torres-Moreno \\
   Laboratoire Informatique d'Avignon - Université d'Avignon  and\\
   Polytechnique Montréal\\
   Avignon Cedex 9, France \\
   \texttt{juan-manuel.torres@univ-avignon.fr} \\
}

\begin{document} 

\maketitle

\begin{abstract}
In this article, we describe some discursive segmentation methods as well as an evaluation of the segmentation quality.
Although our experiment were carried for documents in French,
we have developed three discursive segmentation models solely based on resources simultaneously available in several languages:  marker lists and a statistic POS labeling.
We have also carried out automatic evaluations of these systems against the {\sc Annodis} corpus, which is a manually annotated reference. The results obtained are very encouraging.
\end{abstract}

\keywords{Discourse Segmentation;
Rhetorical Structure Theory;
Multilingual Discourse Parsing.
}     

\section{Introduction}

\begin{figure}[ht]
    \centering
    \includegraphics[width=13cm]{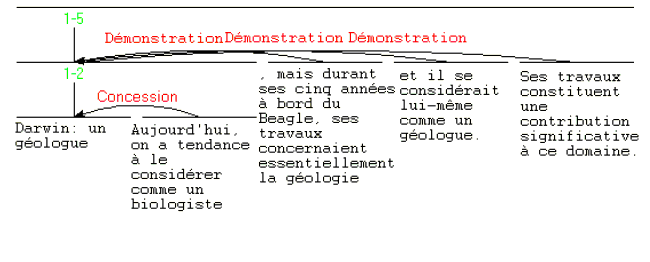}
    \caption{A Rhetorical Structure Theory Tree of a document in French.}
    \label{fig:Fig1}
  \end{figure}

Rhetorical Structure Theory (RST) \cite{manthompson88} is a technique of Natural Language Processing (NLP), in which a document can be structured hierarchically according to its discourse. The generated hierarchy, a tree, provides information associated with the boundaries of the discourse segments and related to their importance and dependencies. The figure \ref{fig:Fig1} shows an example of such a rethorical tree. In the rethorical parsing process, the text has been divided into five units. In the figure \ref{fig:Fig1}, the arrow that leaves the unit (2) towards the unit (1) symbolizes that the unit (2) is the satellite of the unit (1), which is the core in a ``Concession'' relationship. In turn, the units (1) and (2) comprise the nucleus of three ``Demonstration'' relationships.

The discursive analysis of a document normally includes three consecutive steps: 1) discursive segmentation; 2) detection of the discursive relations; 3) construction of the hierarchical rhetorical tree. Regarding the discursive segmentation, there are segmenters in several languages. However, each piece depends on sofisticated linguistic resources, which complicates the reproduction of the experiments in other languages. Consequently, the development of multilingual systems using discursive analysis are yet to be developed. Diverse applications based on the latest technologies require at least one of the three steps mentioned above \cite{molina2013discursive,molinalinguamatica10,compression}. In this context, the idea of exploring the architecture of a generic system that is able not only of segmenting a text correctly but also of adapting it to any language, was a great motivation of this research work.

In this article we show the preliminary results of a generic segmenter composed of several systems (different segmentation strategies). In addition, we describe an automatic evaluation protocol of discursive segmentation. 
The article is composed by the following sections: state of the art (\ref{sec:etat_de_l'art}), which presents a brief bibliographic review; Description of the {\sc Annodis} (\ref{sec:annodis}) corpus \cite{Afantenos:12} used in our tests and of the general architecture of the proposed systems (\ref{sec:protocole}); Segmentation strategies (\ref{sec:StrategieSistema}), which characterizes the different methods implemented to segment the text; results of our numerical experiments ({sec:experiments}); and we conclude with our conclusions and perspectives (\ref{sec:conclusion}).

\section{State-of-the-art} 
\label{sec:etat_de_l'art}

In RST, there are tow discursive units: nuclei and satellites. The nucleus provide information pertinent to the purposes of the author of the text and the satellites add additional information to the nucleu, on which they are dependent on.
In the context of RST, possible discursive relationships may be nucleus-satellite and multinuclear. In nucleus-satellite relationships, a satellite depends on one nucleus, whereas in multinuclear relationships, several nuclei (at least two) are regrouped at the same level of importance (tree hierarchy). Thus, in the discursive segmentation proposes to reduce the text into the minimal discursive units called  Elementary Discursive Units (EDU), through the use of explicit discursive markers. As an example, we can quote some markers in French: 

{\it afin de, pour que, donc, quand bien même que, ensuite, de fois que, globativamente, par contre, sinon, à ce moment-là, cependant, subséquemment, puisque, au fur et à mesure que, si, finalement, etc.}\footnote{so that; so that; therefore; even though; then; times that; globally; on the other hand; otherwise; at that time; however; subsequently; since; as and when; if; finally, etc.;}. 

Markers or particles are often used to connect ideas. 
%Consideremos a frase abaixo: 
Let's consider the sentence below:

%  {\it La ville d'Avignon est \textbf{considérée comme} une belle ville}\footnote{Traduç\~ao da frase: A cidade de Avignon é \textbf{considerada como} uma bela cidade}.

{\it La ville d'Avignon est la capitale du Vaucluse,  \textbf{qui} est un département du sud de la France.}\footnote{Translation of the sentence: The city of Avignon is the capital of Vaucluse \textbf{which} is a department in the south of France.}

\noindent  \textbf{\textit{qui}} (which) is a discursive marker because it connects two ideas. The first one, ``Avignon City is the capital of Vaucluse'' ({\it La ville d'Avignon est la capitale du Vaucluse}), and the second one (satellite), ``[Vaucluse] is a department in the south of France'' ({\it [Vaucluse] est un département du sud de la France}).
Several research has addressed automatic segmentation in several languages, such as: French \cite{afantenos10}, English \cite{tofiloski09}, Portuguese \cite{mazeiro07}, Spanish \cite{da2012diseg,maziero2011dizer} and Tahi. \cite{ketui2012rule}.
 All converge to the idea of using an explicit list of marks in order to segment texts.

Recently, shared task multilingue du workshop Disrpt 2019 was proposed.
In this task, the best system for French segmentation was ToNy \cite{muller-tony}.

\section{{\bf Annodis} Corpus} \label{sec:annodis}

In this exploratory work, our tests considered only documents in French from the {\sc Annodis}\footnote{\url{http://w3.erss.univ-tlse2.fr:8080/index.jsp?perso=annodis&subURL=}.} corpus  \cite{Afantenos:12}.
Annodis ({\it ANNOtation DIScursive}) is a  set of documents in French that were manually enriched with notes of discursive structures. 
Its main characteristics are:

\begin{itemize}
    \item [\textbullet] Two annotations: Rhetorical relations \footnote{\url{http://redac.univ-tlse2.fr/corpus/annodis/annodis_rr.html}} and multilevel structures.
	\item [\textbullet] Documents (687 000 words) taken from four sources:  the {\it Est Républicain} newspaper (39 articles, 10 000 words); Wikipedia (30 articles + 30 summaries, 242 000 words); Proceedings of the conference \textit{Traitement Automatique des Langues Naturelles  (TALN)}\footnote{International French NLP congress.} 2008 (25 articles, 169 000 words); 
Reports from {\it Institut Français de Relations Internationales} (32 raports, 266 000 words).
	\item [\textbullet] The corpora were noted using Glozz.
\end{itemize}

{\sc Annodis}  aims at building an annotated corpus. The proposed annotations are on two levels of analysis, that is, two perspectives:

\begin{itemize}
	\item [\textbullet] Ascendant: part of EDU are used in the construction of more complex structures, through the relations of discourse;
	\item [\textbullet] Descending: approaches the text in its entirety and relies on the various shallow indices to identify high-level discursive structures (macro structures).
 \end{itemize}

Two types of persons annotated {\sc Annodis}: linguistic experts and students. The first group constituted a $E$ subcorpus called ``specialist'' and the second group resulted in a $N$ subcorpus called ``naive''. These rhetorically annotated subcorps were used as references in our experiments. (c.f. \S \ref{sec:experiences}).
%We have almost chosen to use learning methods in order to study the segmentation parameters, but we know that these methods require large amounts of learning data. 
%However, {\sc Annodis}  is small. In addition, the task of rhetorical segmentation is not simple, which compromises the use of efficient learning methods. Thus, we restrict ourselves to follow a strategy of detecting simpler segments, yet quite reproducible.

\section{Discourse Segmenter Overall Description} \label{sec:protocole}

%A figura \ref{fig:archi} mostra a arquitetura geral que serve como base aos sistemas desenvolvidos, onde cada estratégia de segmentaç\~ao configura um novo sistema. Uma das listas do banco de marcadores expl\'icitos é lida pelo segmentador, de acordo com o valor de um parâmetro do sistema, no momento em que o script associado à segmentaç\~ao é executado. Atualmente, dispomos de listas de marcadores em francês, espanhol, inglês e português. Utilizamos a lista do projeto Lexiconn \cite{roze2012lexconn}, que regrupa 328 marcadores da l\'ingua francesa. Um outro parâmetro especifica qual estratégia de segmentaç\~ao deve ser aplicada, segundo a etiquetagem morfo-sint\'atica POS ({\it Part Of Speech}) do documento. 

\begin{figure}[ht]
    \centering
    \includegraphics[width=11cm]{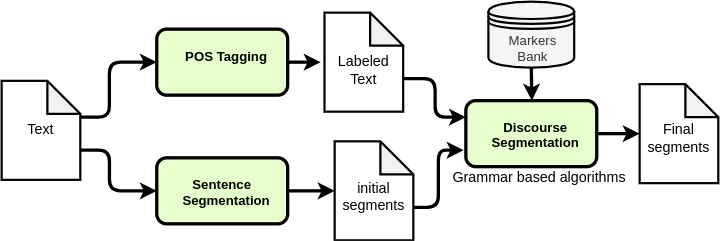}
    \caption{System Architecture Diagram of the proposed Discourse Segmenter.}
    \label{fig:archi}
  \end{figure}

The Figure \ref{fig:archi} shows the general architecture of the proposed discourse segmenter system. The initial input is the raw text encoded in UTF-8. The two initial processes are Part of Speech morphosyntactic Tagging (POS) and the segmentation at the level of the sentences. This last is just a preprocessing step that splits sentences. In the last process the system uses a bank of explicit markers in roder to apply the rules for the final discourse segmentation.

For the experiments, we used lists of markers in French, Spanish, English and Portuguese. We also used the Lexiconn \cite{roze2012lexconn} project list, which regroups 328 French-language markers. Another important parameter specifies which segmentation strategy should be applied, according to the POS labelling of the document. 

\section{Description of segmentation strategies}
\label{sec:StrategieSistema}

\subsection{Segmentation with explicit use of a marker}

%O sistema elementar {\sc Segmentador}$_{\mu}$ ({\it baseline}) se apoia unicamente numa lista de marcadores discursivos para efetuar a segmentaç\~ao. Ele substitui a apariç\~ao de um marcador da lista por um s\'imbolo especial, por exemplo $\mu$, que indica uma fronteira entre o segmento direito e o esquerdo. Seja a frase do exemplo precedente: {\it La ville d'Avignon est la capitale du Vaucluse,  \textbf{qui} est un département du sud de la France.}, O {\sc Segmentador}$_{\mu}$ segmenta a frase em duas partes: o segmento esquerdo (SE), \textit{La ville d'Avignon est la capitale du Vaucluse}, e o direito (SD), \textit{est un département du sud de la France.}.

The elementary system {\sc Segmenter}$_{\mu}$ (baseline) relies solely on a list of discursive markers to perform the segmentation. It replaces the appearance of a marker in the list with a special symbol, for example $\mu$, which indicates a boundary between the right and left segment. Be the sentence of the preceding example: {\it La ville d'Avignon est la capitale du Vaucluse, \textbf{qui} est un département du Sud de la France.}. The Segmenter split the sentence in two parts: the left segment (SE), \textit{La ville d'Avignon est la capitale du Vaucluse}, and the right segment (SD), \textit{est un département du sud de la France}.

\subsection{Segmentation with explicit use of a marker and POS labels}

%O sistema {\sc Segmentador}$_{\mu+}$ apresenta uma melhoria ao {\sc Segmentador}$_{\mu}$: inclus\~ao das categorias gramaticais com a ferramenta {\it TreeTagger}. A vantagem desse sistema consiste na detecç\~ao de certas formas gramaticais a fim de condicionar a segmentaç\~ao. Como ele é baseado no {\sc Segmentador}$_{\mu}$, tentamos reconhecer as condiç\~oes oportunas para reunir dois segmentos quando ambos fazem parte do mesmo segmento discursivo. Buscamos identificar mais sutilmente quando é pertinente deixar os dois segmentos separados. O {\sc Segmentador}$_{\mu+}$ é proposto a partir de duas estratégias diferentes:

The {\sc Segmenter}$_{\mu+}$ system presents an improvement to the {\sc Segmenter}$_{\mu}$: inclusion of grammar categories with the  TreeTagger tool. The advantage of this system is the detection of certain grammatical forms in order to condition the segmentation.
Since it is based on the {\sc Segmenter}$_{\mu}$, we try to recognise the opportune conditions to gather two segments when both are part of the same discursive segment. We try to identify more subtly when it is pertinent to leave the two segments separate. The {\sc Segmenter}$_{\mu}$ has two distinct strategies:

\begin{itemize}
%    	\item [\textbullet] {\sc Segmentador}$_{\mu+V}$ (vers\~ao verbal, V): se apoia unicamente na presença de formas verbais à direita e à esquerda do marcador discursivo. As duas regras grama\-ticais dessa estratégia s\~ao:
    	\item [\textbullet] {\sc Segmentador}$_{\mu+V}$ (verbal version, V): it relies solely on the presence of verbal forms to the right and left of the discursive marker. The two grammatical rules of this strategy are:
		\begin{enumerate}
%	 		\item Se n\~ao existem verbos nos segmentos esquerdo e direito, reagrup\'a-los.
%	  		\item Se existe ao menos um verbo no segmento esquerdo ou direito, os segmentos permanecer\~ao separados.
	 		\item If there are no verbs in the left and right segments, regroup them.
	  		\item If there is at least one verb in the left or right segment, the segments will remain separate.
	  	\end{enumerate}
%	\item [\textbullet] {\sc Segmentador}$_{\mu+(V-N)}$ (vers\~ao verbo-nominal, V-N): se apoia na presença de verbos e de substantivos. Para essa vers\~ao, quatro regras s\~ao consideradas:
	\item [\textbullet] {\sc Segmenter}$_{\mu+(V-N)}$ (verb-noun version, V-N): it relies on the presence of verbs and nouns. For this version, four rules are considered:
	\begin{enumerate}
%	  \item Se n\~ao existe substantivo no segmento esquerdo nem no direito, reagrupamos os segmentos.
%	  \item Reagrupamos os segmentos se ao menos um deles n\~ao possui substantivo. 
%	  \item Se ao menos um substantivo est\'a presente nos dois segmentos, eles permanecem independentes.
%	  \item Se n\~ao existe forma verbo-nominal, os segmentos restam independentes. 
	  \item If there is no noun in either the left or right segment, we regroup the segments.
	  \item We regroup the segments if at least one of them has no noun. 
	  \item If at least one noun is present in both segments, they remain independent.
	  \item If there is no verb-nominal form, the segments remain independent.	  
	  
	\end{enumerate}
\end{itemize}

\section{Experiments}
\label{sec:experiences}

In this first exploratory work, only documents in French were considered, but the system can be adapted to other languages. The evaluation is based on the correspondence of word pairs representing a border. In this way we compare the {\sc Annodis} segmentation with the automatically produced segmentation. 
For each pair of reference segments, a $L_r$ list of word pairs is provided: the last word of the first segment and the first word of the second. 

For example, considering the reference text {\texttt{wik1\_01\_02-04-2006.seg}}, from  {\sc Annodis} corpus:

{\it \noindent [Le Ban Amendment]$_1$ 
[Après avoir adopté la Convention,]\_2 
[un certain nombre de PED et d'associations de défense de l'environnement soutinrent]\_3 
[que le document n'allait pas assez loin.]\_4 [De nombreux pays et ONG militèrent]\_5
[en faveur d'une interdiction totale de l'expédition de déchets dangereux à destinations des PED.]\_6 
[Plus exactement,]\_7 [la Convention originale n'interdisait pas l'exportation de déchets,]\_8
[excepté vers l'Antarctique.]\_9 [Elle n'exigeait]\_10 [qu'une procédure de consentement préalable en connaissance de cause]\_11
[(PIC, Prior Informed Consent).]\_12
}

\noindent Here are the word pairs of the created reference list (punctuation marks are disregarded):

$L_r$=\{{\it
[Convention -- un],
[soutinrent -- que],
[loin -- de],
[militèrent -- en],
[exactement -- la],
[PED -- plus],
[exactement -- la],
[déchets -- excepté],
[Antartique -- Elle],
[exigeait -- qu'une],
[cause -- PIC]}
\}

We decided to count the word pairs instead of the segments, as this is a first version of the evaluation protocol. In fact, the segments may be nested, which complicates the evaluation process. Although there are some errors, word boundaries allow us to detect segments more easily.

We have built a second $L_c$ list for the automatically identified segments, following the same criteria of $L_r$. The $L_r$ and $L_c$ lists regroup, pair by pair, the segment border. We then count the common pair intersection of the two lists. Each pair in the $L_c$ list is also present in the $L_r$ reference list and is a {correctly assigned to the class} pair. A word pair belonging to the $L_c$ list but not belonging to the $L_r$ reference list, will be a pair \textbf{assigned to the class}.
\noindent For that same text, the $L_c$ list of candidate pairs obtained with the {\sc Segmentator}$_{\mu}$ is:

$L_r$=\{{\it
[loin--De],
[pays--et],
[militèrent--en],
[dangereux--à],
[PED--Plus],
[Antarctique--Elle],
[préalable--en],
[cause--PIC]}
\}

We calculate the precision $P$, the recall $R$ and the $F$-score on the text corpus used in our tests, as follow: 

\begin{eqnarray}
P &=& (\textrm{Nb of pairs $\in L_c$} \cap \textrm{pairs $\in L_r$}) / \textrm{Nb of pairs $\in L_c$}\\
R &=& (\textrm{Nb of pairs $\in L_c$} \cap \textrm{pairs $\in L_r$}) / \textrm{Nb of pares $\in L_r$}\\
F\textrm{-score} &=& 2 \times \frac{ P \cdot R  }{ P + R  } 
\end{eqnarray}

\noindent The precision, the recall and the $F$-score for this example is:
$P$ = 5 / 11 = 0.45; $R$ = 5 / 8 = 0.625; F-score = 2 $\times  \frac{ 0.45 \times 0.625}{ 0.45 + 0.625} = 0.523$. 
We used the documents in the {\sc Annodis} corpus without segmentation, because they had been segmented with the {\sc Segmenter}$_{\mu}$  and with the grammar segmenters.

Two batch of tests were performed. The first on the $D$ set of documents common to the two subcorpus ``specialist'' $E$ and ``naive'' $N$ from {\sc Annodis}. $D$ contains 38 documents with 13 364 words. This first test allowed to measure the distance between the human markers. In fact, in order to get an idea of the quality of the human segmentations, the cuts in the texts made by the specialists were measured {it versus} the so-called ``naifs'' note takers and vice versa. The second series of tests consisted of using all the documents of the subcorpus ``specialist'' $E$, because the documents of the subcorpus of {\sc Annodis} are not identical. Then we benchmarked the performance of the three systems automatically.

\subsection{Results}

In this section we will compare the results of the different segmentation systems through automatic evaluations. 
Firstly, the human segmentation, from the subcorpus $D$ composed of common documents. The results are presented in the table \ref{tab:humains}. The first row shows the performance of the $I$ segments, taking the experts as a reference, while the second presents the process in the opposite direction.

\begin{table}[h!]
\center
\begin{tabular}{r|ccc}
\hline
  \bf Reference & \bf F-score & $P$ & $R$ \\
\hline
   Expert ($E$)   & 0.961 &\bf 0.984 & 0.941 \\
   Naive ($N$)    & 0.961 & 0.972 &\bf 0.952\\

\hline
\end{tabular}
\caption{Performance of human segmentations}
\label{tab:humains}
\end{table}

%N\'os constatamos que a segmentaç\~ao realizada pelos especialistas e ingênuos produz dois subcorpus $E$ e $I$ com caracter\'isticas muito similares. Isso nos surpreendeu, pois esper\'avamos uma diferença mais importante entre eles. De toda forma, deduzimos que, ao menos nesse corpus, n\~ao é necess\'ario ser um especialista em lingu\'istica para segmentar discursivamente os documentos. No que concerne às avaliaç\~oes dos sistemas, utilizamos os 78 documentos de $E$ como referência. A tabela \ref{tab:systemes} exibe os resultados obtidos.

We have found that segmentation by experts and naive produces two subcorpus $E$ and $N$ with very similar characteristics. 
This surprised us, as we expected a more important difference between them. In any case, we deduced that, at least in this corpus, it is not necessary to be an expert in linguistics to discursively segment the documents. As far as system evaluations are concerned, we use the 78 $E$ documents as reference. Table \ref{tab:systems} shows the results.

%\begin{table}[h!]
%\center
%\begin{tabular}{r|ccc|ccc}
%\hline
%  & \multicolumn{3}{|c|}{\bf Especialista} & \multicolumn{3}{|c}{\bf Ingênuo}  \\
%  \bf Sistema & \bf F-escore & $P$ & $R$ & \bf F-escore & $P$ & $R$ \\
%\hline
%%   {\sc Segmentador}$_{\mu}$            &\bf 0.5150 & \bf 0.6490 & 0.4349 & \bf 0,5034 & \bf 0,6150 & 0,4360 \\
%%   Gramatical (V)   			& 0.5109 & 0.6092 &\bf 0.4478 & 0,4880 & 0,5725 & 0,4355 \\
%%   Gramatical (V-N) 			& 0.5036 & 0.6107 & 0.4366 & 0,49587 & 0,5730 &\bf 0,4482\\
%   {\sc Segmentador}$_{\mu}$            &\bf 0.4006 & \bf 0.3748 & 0.4505 & \bf 0,5034 & \bf 0,6150 & 0,4360 \\
%   Gramatical (V)   			& 0.5109 & 0.6092 &\bf 0.4478 & 0,4880 & 0,5725 & 0,4355 \\
%   Gramatical (V-N) 			& 0.5036 & 0.6107 & 0.4366 & 0,49587 & 0,5730 &\bf 0,4482\\
%\hline
%\end{tabular}
%\caption{Desempenho dos segmentadores autom\'aticos} % versus Especialista}
%\label{tab:systemes}
%\end{table}

%
\begin{table}[h!]
\center
\begin{tabular}{r|ccc}
\hline
  \bf System & \bf F-score & $P$ & $R$ \\
\hline
   {\sc Segmenter}$_{\mu}$      & 0.416 & 0.388 & \bf 0.463 \\
   Gramatical (V)   			& 0.493 & \bf 0.614 & 0.420 \\
   Gramatical (V-N) 			& \bf 0.494 & 0.594 & 0.431 \\
\hline
\end{tabular}
\caption{Performance of Automatic Segmenters vs. Expert}
\label{tab:systems}
\end{table}

%No caso dos Especialistas, a vers\~ao gramatical verbo-nominal (V-N) mostrou um melhor desempenho F-escore. 
In the case of the Experts, the grammatical verb-nominal version (V-N) had better F-score performance. 
%A vers\~ao verbal (V) obteve uma melhor precis\~ao $P$ que a verbo-nominal (V-N). No caso dos Ingênuos, le desempenho F-escore, $P$ e $R$ e muito similar dos Especialistas.
The verbal version (V) obtained a better accuracy $P$ than the verb-nominal (V-N). 
In the case of the Naive, the performance F-score, $P$ and $R$ is very similar from the Experts.

\section{Conclusions and perspectives} 
\label{sec:conclusion}

%O objetivo deste trabalho era duplo: conceber um segmentador discursivo {\it baseline} utilizando um m\'inimo de recursos e estabelecer um protocolo de avaliaç\~ao objetivo para medir o desempenho dos segmentadores. Os resultados mostram que podemos construir uma vers\~ao {\it baseline} simples, que emprega unicamente uma lista de marcadores, apresentando um desempenho muito encorajador. Evidentemente, a qualidade da lista é um fator prepoderante para uma segmentaç\~ao correta.

The aim of this work was twofold: to design a discursive segmenter using a minimum of resources and to establish an evaluation protocol to measure the performance of segmenters. The results show that we can build a simple version of the baseline, which employs only a list of markers and presents a very encouraging performance. Of course, the quality of the list is a preponderant factor for a correct segmentation.

%N\'os estudamos o impacto do marcador v\'irgula o qual, mesmo parecendo fr\'agil, contribui à melhoria do desempenho dos nossos segmentadores. Assim, trata-se de um marcador interessante que podemos considerar como um marcador discursivo. A vers\~ao {\sc Segmentador}$_{\mu}$ fornece os melhores resultados em termos do F-escore e {\sl recall}, seguido da vers\~ao {\sc Segmentador}$_{\mu+V}$ , que a depassa em precis\~ao. No tocante à avaliaç\~ao, desenvolvemos um protocolo simples que permite comparar o desempenho dos sistemas. Trata-se, a nosso conhecimento, da primeira avaliaç\~ao autom\'atica em francês.

We have studied the impact of the marker which, even though it may seem fringe-worthy, contributes to improving the performance of our segmenters. Thus, it is an interesting marker that we can consider as a discursive marker. The Segmentator$_{\mu}$ version provides the best results in terms of F-score and recall, followed by the Segmentator$_{\mu+V}$ version, which passes it in precision. Regarding evaluation, we developed a simple protocol to compare the performance of the systems. 
%This is, to our knowledge, the first automatic evaluation in French.

%\'E necess\'ario intensificar nossas pesquisas a fim de propor melhorias aos nossos segmentadores, bem como estudar mais aprofundadamente o impacto das regras das etiquetas gramaticais na segmentaç\~ao. Visto que dispomos de um protocolo de avaliaç\~ao padr\~ao, pretendemos realizar testes com o português, o espanhol (ver \cite{da2011development}), o inglês, etc. Para isso, necessitaremos unicamente de uma lista de marcadores de cada l\'ingua.

It is necessary to intensify our research in order to propose improvements to our segmenters, as well as to study further the impact of grammar tag rules on segmentation. Since we have a standard evaluation protocol, we intend to carry out tests with Portuguese, Spanish (see \cite{da2011development}), English, etc. For that, we will only need a list of markers for each language.

%Les performances des systèmes restent, bien entendu modestes, mais il ne faut pas oublier qu'il s'agit de baselines, dont leur vocation primaire est de constituer des systèmes-étalon utilisables dans de protocoles de tests, comme celui que nous avons proposé. Malgré ces performances, ces baselines (ou leurs améliorations) peuvent être utilisées dans des applications comme le résumé automatique de documents par exemple \cite{favre2006lia}, ou la compression de phrases \cite{molina2011discourse}. 

%O desempenho dos sistemas resta modesto, é claro, mas não podemos esqueçer que se trata de uma baseline e seu objetivo prim\'ario é fornecer sistemas-padrão que possam ser utilizados em protocolos de testes, como o que propusemos. Apesar dessa evolução, essas baselines (ou suas vers\~oes melhoradas) podem ser utilizadas em aplicações tais como a sumarização automática de documentos (por exemplo, \cite{favre2006lia}), ou compressão de phrases \cite{molina2011discourse}.

The performance of the systems remains modest, of course, but we must not forget that this is a baseline and its primary objective is to provide standard systems that can be used in testing protocols such as the one we proposed. Despite this evolution, these baselines (or their improved versions) can be used in applications such as automatic document summarisation (e.g., \cite{Torres2014,favre2006lia}), or sentences compression  \cite{molina2011discourse}.

%Le système baselines proposés ont comme caractéristiques principales le fait d'être faiblement couplés à chaque langue.
%En effet, ils utilisent uniquement une liste de marqueurs linguistiques et la catégorie grammaticale des mots.
%La prémière ressource, bien que dependante de chaque langue reste relativement facile à obtenir.
%Nous avons constaté que, même avec des listes de taille très modeste, les resultats sont tout à fait honorables.
%Les catégories grammaticales ont été obtenues à l'aide de l'outil statistique TreeTagger, mais il pourrait être substitué par n'importe quel autre outil produisant des résultats similaires.

%O sistema baseline proposto tem como principal característica sua flexibilidade come relaç\~ao ao idioma considerado.
%Na verdade, ele só usa uma lista de marcadores linguísticos e a categoria gramatical das palavras.
%O primeiro recurso, embora dependente de cada língua, é relativamente fácil de obter.
%Descobrimos que, mesmo com listas de tamanho moderado, os resultados são bastante significativos.
%As categorias gramaticais foram obtidas com a ajuda da ferramenta estat\'istica {\it TreeTagger}. Contudo, {\it TreeTagger} poderia ser substituído por qualquer outra ferramenta produzindo resultados similares.
The main feature of the proposed baseline system is its flexibility with respect to the language considered.
In fact, it only uses a list of language markers and the grammatical category of words.
The first resource, although dependent on each language, is relatively easy to obtain.
We have found that, even with lists of moderate size, the results are quite significant.
The grammatical categories were obtained with the help of the TreeTagger statistics tool. However, {TreeTagger} could be replaced by any other tool producing similar results.

\section*{Appendix}

%Neste anexo, apresentamos a lista de conectores ret\'oricos em francês que constitue nossa lista de marcadores. 
%Ressaltamos que os marcadores terminando em ap\'ostrofo
%(tais como
%s\~ao suprimidos de uma express\~ao regular que implique em

In this appendix, we present the list of rhetorical connectors in French that constitute our list of markers.
We point out that the markers ending in apostrophe such as:

{\it près qu', à condition d',} etc.

are deleted from a regular expression implying 
'and': {\it près qu' + près que, à condition d' + à condition de}, etc.

\vspace{0.3cm}

\hrule

\begin{multicols}{3}
\noindent

, /
à /
à ça près qu' /
à ceci près qu' /
à cela près qu' /
à ce moment-là /
à ce point qu' /
à ce propos /
à cet égard /
à condition d' /
à condition qu' /
à défaut d' /
à défaut de /
à dire vrai /
à élaborer /
à en /
afin d' /
afin qu' /
afin que /
à force /
à force d' /
ainsi /
à la place /
à la réflexion /
à l'époque où /
à l'heure où /
à l'instant où /
à l'inverse /
alors /
alors même qu' /
alors qu' /
à mesure qu' /
à moins d' /
à moins qu' /
à part ça /
à partir du moment où /
à part qu' /
après /
à présent qu' /
après qu' /
après quoi /
après tout /
à preuve /
à propos /
à seule fin d' /
à seule fin qu' /
à supposer qu' /
à telle enseigne qu' /
à tel point qu' /
attendu qu' /
au bout du compte /
au cas où /
au contraire /
au fait /
au fur et à mesure qu' /
au lieu /
au lieu d' /
au même titre qu' /
au moins /
au moment d' /
au moment où
auparavant /
au point d' /
au point qu' /
aussi /
aussi longtemps qu' /
aussitôt /
aussitôt qu' /
autant /
autant dire qu' /
au total /
autrement /
autrement dit /
avant /
avant d' /
avant même d' /
avant même qu' /
avant qu' /
à vrai dire /
bien qu' /
bientôt /
bref /
car /
ceci dit /
ceci étant dit /
cela dit /
cependant /
cependant qu' /
c'est à dire qu' /
c'est pourquoi /
cette fois qu' /
comme /
comme ça /
comme quoi /
comme si /
comparativement /
conséquemment /
considérant qu' /
considéré qu' /
corrélativement /
d'abord /
d'ailleurs /
dans ce cas /
dans ce cas-là /
dans la mesure où /
dans le but d' /
dans le but qu' /
dans le cas où
dans le coup /
dans le sens où /
dans le sens qu' /
dans l'espoir d' /
dans l'espoir qu' /
dans l'hypothèse où /
dans l'intention d' /
dans l'intention qu' /
dans tous les cas /
d'autant plus qu' /
d'autant qu' /
d'autre part /
de ce fait /
décidément /
de façon à /
de façon à ce qu' /
de façon qu' /
de fait /
déjà /
déjà qu' /
de la même façon /
de la même façon qu' /
de la même manière /
de la même manière qu' /
de manière à /
de manière à ce qu' /
de manière qu' /
de même /
de même qu' /
de plus /
depuis /
depuis qu' /
des fois qu' /
dès lors /
dès lors qu' /
de sorte qu' /
dès qu' /
de telle façon qu' /
de telle manière qu' /
de toute façon /
de toute manière /
de toutes façons /
de toutes manières /
d'ici qu' /
dire qu' /
donc /
d'où /
d'où qu' /
du coup /
du fait qu' /
du moins /
du moment qu' /
d'un autre côté
d'un côté /
d'un coup /
d'une part /
d'un seul coup /
du reste /
du temps où /
effectivement /
également /
en /
en admettant qu' /
en attendant /
en bref /
en ce cas /
en ce sens qu' /
en comparaison /
en conséquence /
encore /
encore qu' /
en d'autres termes /
en définitive /
en dépit du fait qu' /
en dépit qu' /
en effet /
en fait /
enfin /
en gros /
en même temps /
en même temps qu' /
en outre /
en particulier /
en plus /
en plus d' /
en plus de /
en réalité /
en résumé /
en revanche /
en somme /
ensuite /
en supposant qu' /
en tous cas
en tous les cas /
en tout cas /
en tout état de cause /
en vérité /
en vue d' /
et /
étant donné qu' /
et dire qu' /
et puis /
excepté qu' /
faute d' /
finalement /
globalement /
histoire d' /
hormis le fait qu' /
hormis qu' /
instantanément /
inversement /
jusqu'à /
jusqu'à ce qu' /
la preuve /
le fait est qu' /
le jour où /
le temps qu' /
lorsqu' /
maintenant /
maintenant qu' /
mais /
malgré le fait qu' /
malgré qu' /
malgré tout /
malheureusement /
même /
même qu' /
même si /
mieux /
mis à part le fait qu' /
mis à part qu' /
néanmoins /
nonobstant /
nonobstant qu' /
or /
ou /
ou bien /
outre qu' /
par ailleurs /
parallèlement /
parce qu' /
par comparaison /
par conséquent /
par contre /
par-dessus tout /
par exemple /
par le fait qu' /
par suite /
pendant qu' /
peu importe
plus qu' /
plus tard
plutôt /
plutôt qu' /
plutôt que d' /
pour /
pour autant
pour autant qu' /
pour commencer /
pour conclure /
pour finir /
pour le coup /
pour peu qu' /
pour preuve /
pour qu' /
pour résumer /
pourtant /
pour terminer /
pour une fois qu' /
pourvu qu' /
premièrement /
preuve qu' /
puis /
puisqu' /
quand /
quand bien même /
quand bien même qu' /
quand même /
quant à /
quitte à /
quitte à ce qu' /
quoiqu' /
quoi qu'il en soit /
réciproquement /
réflexion faite /
remarque /
résultat /
s' /
sachant qu' /
sans /
sans compter qu' /
sans oublier qu' /
sans qu' /
sauf à /
sauf qu' /
selon qu' /
si /
si bien qu' /
si ce n'est qu' /
simultanément /
sinon /
sinon qu' /
si tant est qu' /
sitôt qu' /
soit /
soit dit en passant /
somme toute /
soudain /
subséquemment /
suivant qu' /
surtout /
surtout qu' /
tandis qu' /
tant et si bien qu' /
tant qu' /
total /
tout à coup /
tout au moins /
tout bien considéré /
tout compte fait /
tout d'abord /
tout de même /
tout en /
une fois qu' /
un jour /
un jour qu' /
un peu plus tard /
vu qu' /
\end{multicols}

\newpage

\bibliographystyle{plain}
\bibliography{biblio}

\end{document}